# Q-Gaussian Swarm Quantum Particle Intelligence on Predicting Global Minimum of Potential Energy Function

Hiqmet Kamberaj[1]

Faculty of Technical Sciences, Department of Information Technology, International Balkan University, Skopje, R. of Macedonia

*Abstract*

We present a newly developed $q$-Gaussian Swarm Quantum-like Particle Optimization (q-GSQPO) algorithm to determine the global minimum of the potential energy function. Swarm Quantum-like Particle Optimization (SQPO) algorithms have been derived using different attractive potential fields to represent swarm particles moving in a quantum environment, where the one which uses a harmonic oscillator potential as attractive field is considered as an improved version. In this paper, we propose a new SQPO that uses $q$-Gaussian probability density function for the attractive potential field (q-GSQPO) rather than Gaussian one (GSQPO) which corresponds to harmonic potential. The performance of the q-GSQPO is compared against the GSQPO. The new algorithm outperforms the GSQPO on most of the time in convergence to the global optimum by increasing the efficiency of sampling the phase space and avoiding the premature convergence to local minima. Moreover, the computational efforts were comparable for both algorithms. We tested the algorithm to determine the lowest energy configurations of a particle moving in a 2, 5, 10, and 50 dimensional spaces.

---

[1] Corresponding author: h.kamberaj@gmail.com; Tel. +389(0)75462189; Fax. +389(0)23214832;

## I. Introduction

The task of finding the global optimum in a multimodal problem is notoriously difficult since the number of stable optima increases as the search space increases. For example, the search for the global minimum energy in a surface energy landscape of the atomic structures (1). Swarm Particle Optimization (SPO) is a versatile population-based optimization technique, in many respects similar to evolutionary algorithms (2). Kennedy & Eberhart introduced the concept of function-optimization by means of a particle swarm (3). Suppose the global optimum of a $n$ dimensional function is to be located. The function may be mathematically represented as

$$f(x_1, x_2, x_3, \ldots, x_n) = f(\vec{X}),$$

where $\vec{X}$ is the search-variable vector, which actually represents the set of independent variables of the given function. The task is to find out such a $\vec{X}^*$, that the function value $f(\vec{X}^*)$ is a minimum, denoted by $f^*$, in the search range.

The SPO has its origin on the swarm intelligence algorithms, which are concerned with the design of intelligent multi-agent systems by taking stimulation from the collective behavior of social insects such as ants, termites, bees, and wasps, as well as from other animal societies such as flocks of birds or schools of fish (3). In SPO method, the particles that represent potential solutions move around in the phase space with a velocity updated by the particle's own experience and the experience of the particle's neighbors or the experience of the whole swarm. SPO has been shown to perform well for many problems (4). Recently this technique has also shown interests in statistical physics. (5)

Many variants of the basic algorithm and the applications of SPO have been proposed. (6; 7; 8) Since the standard SPO has a low convergence rate (9), a new approach has been proposed based on the quantum mechanics and the delta potential well model to sample around the previous best points (10; 11), named swarm quantum particle optimization (SQPO), which is considered as a probabilistic method. (12) Other variants of SQPO methods have been implemented (13). Several improvements have been introduced for SQPO as described in this review. (14) These improvements are mainly focused on how to improve parameter selection (15; 16) or maintaining diversity of the swarm (17; 18; 19). The use of the harmonic potential well is considered to be one of these improvements, for which the probability of finding the swarm quantum-like particle at a certain position in the phase space is described by Gaussian distribution. (20)

In recent years (21; 22; 23), it has been shown that the use of probability distributions with heavy tails can be useful in allowing the system to escape from local optima in multimodal problems. Other improvements of the SQPO method have also been shown (24). A review of the SQPO methods is presented in Ref. (25). The use of occupations taken from long tails of the distribution implies jumps of scale-free sizes, eventually allowing reaching distant regions of the search space faster. (21; 23; 26)

This property is interesting for SQPO algorithm too, which will be shown in our study, as it can allow the swarm particles to escape faster from local minima located close to the best solution before the change.

The problem facing GSQPO algorithm is the premature convergence to a local minima due to low diversity of the swarm particles. (27) The main contribution of this paper is the investigation of the use of the $q$-Gaussian distribution for probability distribution of the swarm particles. The Gaussian distribution is ever present in probability and statistics due to its role as an attractor of independent systems with finite variance. It is also the distribution which maximizes the well-known Boltzmann-Gibbs entropy under appropriate constraints (28). The $q$-Gaussian distribution arises as an attractor of certain correlated systems, or when maximizing the so-called Tsallis entropy under appropriate constraints (29). The $q$-Gaussian distributions are ubiquitous within the framework of ``non-extensive statistical mechanics''. It has widely been used in chemistry, engineering, and computational sciences, such as the so called *Generalized Simulated Annealing* (30), in which the visiting steps for possible jump acceptance are determined by $q$-Gaussians instead of the traditional Gaussians. The generalized distributions, such as q-Gaussian, as a probability distribution with long tails will increase the diversity of the swarm allowing the particle swarm reaching long distant regions of space, increasing in this way the efficiency for searching the global minima.

## II. Materials and Methods

*GSQPO algorithm*

The swarm quantum-like particle optimization algorithm allows all swarm particles to move under quantum-mechanical laws (20), that is, we can determine the probability of finding the swarm particle at the position $\vec{X}$ any time $t$. In analogy with the fundamental hypotheses of quantum mechanics, the probability distribution function of swarm particle appearing in a position $\vec{X}$ is determined by $|\psi(\vec{X},t)|^2$, where $\psi(\hat{X},t)$ is the state vector, which depends on the potential field the particle lies in. Based on quantum mechanics, $\psi(\hat{X},t)$ is the solution of the time-dependent Schrödinger equation (31):

$$i\hbar \frac{\partial \psi(\vec{X},t)}{\partial t} = \hat{H}(\vec{X})\psi(\vec{X},t) \tag{1}$$

with $\hbar$ being the Planck's constant, and $i = \sqrt{-1}$. $\hat{H}(\vec{X})$ is a time-independent Hamiltonian operator of the system, defined as

$$\hat{H}(\vec{X}) = -\frac{\hbar^2}{2m}\nabla^2 + U(\vec{X}) \tag{2}$$

where $m$ is a fiction mass of the swarm particle, and $U(\vec{X})$ is the potential energy function of the attractive center.

The design step in deriving the hybrid SQPO algorithm was proposed in (11; 10) is the choice of a suitable attractive potential field that can guarantee bound states for the particles moving in the quantum environment. A potential distribution, which is very common in quantum mechanics, is the harmonic oscillator potential well, which is known to have the following solution

$$\psi_n(\vec{X}) = \left(\frac{\alpha}{2^L L! \pi^{1/2}}\right)^{1/2} H_L(\alpha \vec{X}) e^{-(\alpha X)^2/2} \tag{3}$$

where $\alpha = (mk/\hbar)^{1/4}$ and $H_L$ is the Hermite polynomial (32), and $k$ is the potential well depth. The index $L$ defines multiple possible state vectors of the system. The ground-state solution is obtained for $L = 0$, which is considered to be a simple solution that describes many problems. In this case the probability distribution is a Gaussian distribution given as

$$f(X) = \frac{\alpha}{\sqrt{\pi}} e^{-\alpha^2 X^2} \tag{4}$$

In this context, employing the Monte Carlo method, the particles move according to the following iterative equation (20):

$$\begin{cases} X_{ij}(t+1) = p_j + \dfrac{\beta_t}{\alpha_{t+1}} \cdot F(u) & z \geq 0.5 \\ X_{ij}(t+1) = p_j - \dfrac{\beta_t}{\alpha_{t+1}} \cdot F(u) & z < 0.5 \end{cases} \quad (5)$$

where $i = 1,2,\cdots,N$ indicates a particle of population, and $j = 1,2,\cdots,d$ indicates the dimension of the phase space, and $F(u)$ is a random variable with probability distribution $F$. The design parameter $\beta_t$ is called *contraction–expansion coefficient* (33); $u$ and $z$ are random values generated according to a uniform probability distribution in the range $[0,1]$. The so-called local attractor, $p_j$, is defined as the following to guarantee convergence of the optimization method (3; 34):

$$p_j = RX_{ij}^{LBest} + (1-R)X_j^{GBest}, \quad (6)$$

where $R$ is a random number in $[0,1]$ and $X_j^{GBest}$ is the global best position along the $j$th dimension. Note that $\vec{p}$, given by Eq. (14), is the center of the potential well.

Here, $1/\alpha_t$ is the time-dependent characteristic length. For the standard GSQPO algorithm, we need to enforce a time evolving parameter $1/\alpha_t$ such that all particles will eventually arrive to the desired location. To guarantee, on the average, that the next particle will converge, we need the value of $|r_{t+1}|$ at iteration $(t+1)$ to be closer to zero. The condition of this convergence for the algorithm is given by (20)

$$\lim_{t \to +\infty} \frac{1}{\alpha_t} = 0 \quad (7)$$

A suitable probabilistic translation for this statement is given by (20):

$$P(t \to t+1) = \int_{-\infty}^{|r_t|} f_{t+1}(x)\, dx \quad (8)$$

where $f_{t+1}(x)$ is the probability density function of the particle at the $(t+1)$-th iteration. The value of $\alpha$ is chosen as optimal solution of the following inequality

$$P(t \to t+1) > P_o \qquad (9)$$

where $P_o$ is the smallest transition probability, which guarantees on the average that the next particle will converge. Inequality (9) can be further reduced to the expression

$$\int_0^{|r_t|} f_{t+1}(x)\, dx > P_o - 1/2 \qquad (10)$$

Integration of the integral in Eq. (10) gives a solution in terms of error function (*erf*) as

$$erf(\alpha_{t+1}|r_t|) > 2P_o - 1 \qquad (11)$$

We look for the solution of Eq. (11) of this form

$$\frac{1}{\alpha_{t+1}} = g \frac{|r_t|}{y_o}. \qquad (12)$$

where $y_o$ is the solution of this equation: $erf(y) - (2P_o - 1) = 0$ and $g$ is a scaling factor less than one. Then, Eq. (5) can be re-written as

$$\begin{cases} X_{ij}(t+1) = p_j + g \dfrac{\beta_t}{y_o} \cdot \left| M_j^{Best} - X_{ij}(t) \right| \cdot F(u) & z \geq 0.5 \\ X_{ij}(t+1) = p_j - g \dfrac{\beta_t}{y_o} \cdot \left| M_j^{Best} - X_{ij}(t) \right| \cdot F(u) & z < 0.5 \end{cases} \qquad (13)$$

where $|r_i| = |\vec{M}^{Best} - \vec{X}_i(t)|$, with $\vec{M}^{Best}$ being the global point, also called *Mainstream Thought* or *Mean Best* of the population (24), which is defined as the mean of the $\vec{X}_i^{LBest}$ best positions of all particles ($i = 1, 2, \cdots, N$) in each dimension and it is given by (24)

$$\vec{M}^{Best} = \frac{1}{N} \sum_{i=1}^{N} \vec{X}_i^{LBest}. \tag{14}$$

*q-GSQPO algorithm*

The main motivation of this paper is the investigation of the use of the $q$-Gaussian distribution for probability distribution of the swarm particle. The hope is that the use of probability distributions with longer tails will allow the system escaping from local minima, eventually allowing reaching distant regions of the search space faster. The standard $q$-Gaussian distribution is specified by the probability distribution function

$$f_q(x) = A_0 \left[ 1 - (1-q)\alpha_q^2 x^2 \right]^{1/(1-q)} \tag{15}$$

for $q > 1$. It is normalized if

$$A_0 = \frac{\alpha_q}{\sqrt{\pi}} \sqrt{q-1} \frac{\Gamma\left[\frac{1}{q-1}\right]}{\Gamma\left[\frac{3-q}{2(q-1)}\right]}$$

Recent works have been focused on the study of mathematical properties of $q$-Gaussian functions (35), including methods for generating random numbers which follow $q$-Gaussian distributions (36). In this study, we used the method proposed in Ref. (36) for generating random number according to $q$-Gaussian distribution. The derivation of the corresponding formulas for $q$-Gaussian requires some more efforts due to the difficulties in the integration of Eq. (10). Again, the equations of the algorithm are of the same form as these given by Eq. (5):

$$\begin{cases} X_{ij}(t+1) = p_j + \dfrac{\beta_t^{(q)}}{\alpha_{t+1}^{(q)}} \cdot F_q(u) & z \geq 0.5 \\ X_{ij}(t+1) = p_j - \dfrac{\beta_t^{(q)}}{\alpha_{t+1}^{(q)}} \cdot F_q(u) & z < 0.5 \end{cases} \quad (16)$$

where $F_q(u)$ is now a random number following $q$-Gaussian distribution. The transition probability, calculated by integration of Eq. (8) with probability density function given by Eq. (15), is given in terms of the Hypergeometric functions (32), $_2F_1(a,b;c;z)$ as:

$$P(t \to t+1) = \sqrt{\dfrac{1}{\pi} \dfrac{q-1}{3-q}} \, \alpha_{t+1}^{(q)} |r_t| \dfrac{\Gamma\left(\dfrac{1}{q-1}\right)}{\Gamma\left(-\dfrac{1}{2} + \dfrac{1}{q-1}\right)} \, _2F_1\left(\left[\dfrac{1}{2}, \dfrac{1}{q-1}\right], \dfrac{3}{2}, \dfrac{q-1}{q-3}\left(\alpha_{t+1}^{(q)}|r_t|\right)^2\right) + \dfrac{1}{2} \quad (17)$$

From Eq. (10), we get:

$$\sqrt{\dfrac{1}{\pi} \dfrac{q-1}{3-q}} \, \alpha_{t+1}^{(q)} |r_t| \dfrac{\Gamma\left(\dfrac{1}{q-1}\right)}{\Gamma\left(-\dfrac{1}{2} + \dfrac{1}{q-1}\right)} \, _2F_1\left(\left[\dfrac{1}{2}, \dfrac{1}{q-1}\right], \dfrac{3}{2}, \dfrac{q-1}{q-3}\left(\alpha_{t+1}^{(q)}|r_t|\right)^2\right) > P_\circ - \dfrac{1}{2} \quad (18)$$

As one can see, the solution depends on both, the parameter $q$ and $P_\circ$. Again, we express

$$\dfrac{1}{\alpha_{t+1}^{(q)}} = g \dfrac{|r_t|}{y_\circ^{(q)}} \quad (19)$$

Here, $y_\circ^{(q)}$ is the solution of the following equation:

$$\sqrt{\frac{1}{\pi}\frac{q-1}{3-q}}\,\alpha_{t+1}^{(q)}|r_t|\frac{\Gamma\left(\frac{1}{q-1}\right)}{\Gamma\left(-\frac{1}{2}+\frac{1}{q-1}\right)}\,{}_2F_1\left(\left[\frac{1}{2},\frac{1}{q-1}\right],\frac{3}{2},\frac{q-1}{q-3}\left(\alpha_{t+1}^{(q)}|r_t|\right)^2\right)-\left(P_\circ-\frac{1}{2}\right)=0 \qquad (20)$$

with respect to $\alpha_{t+1}^{(q)}|r_t|$, which is solved numerically for different chosen values of $q$ and $P_\circ$ as discussed in the following section. The constant, $g$, is again scaling factor less than one.

The following equations are proposed to describe the q-GSQPO algorithm:

$$\begin{cases} X_{ij}(t+1) = p_j + g\,\dfrac{\beta_t^{(q)}}{y_\circ^{(q)}}\cdot\left|M_j^{Best} - X_{ij}(t)\right|\cdot F_q(u) & z \geq 0.5 \\[2ex] X_{ij}(t+1) = p_j - g\,\dfrac{\beta_t^{(q)}}{y_\circ^{(q)}}\cdot\left|M_j^{Best} - X_{ij}(t)\right|\cdot F_q(u) & z < 0.5 \end{cases} \qquad (21)$$

*Optimization of the algorithms*

The main problem with standard SQPO applications in optimization problems is that it will eventually converge to an optimum; it thereby loses the diversity necessary for efficient exploration of the search space. There have been proposed other methods for $F(u)$ (14), such as an exponential distribution, however, the Gaussian distribution assumed above for the particle positions is more efficient in global search ability than the exponential distribution (20). In fact different choices of the potential distribution will yield different portability distribution for the swarm dynamics (see also discussion in Ref. (20)). The case of the harmonic oscillator, which corresponds to parabolic potential distribution, gives a Gaussian distribution of the swarm dynamics. The Gaussian behavior can be used to simplify the canonical form of the classical SPO method by using the update of only position according to a Gaussian distribution. The fact that the Gaussian distribution is direct consequence of the solution of the Schrödinger equation, using a power law for potential distribution, opens the possibility for other probability distributions as well with different physical insights. In this study we use the $q$-Gaussian distribution, which is developed within the framework of non-extensive statistical mechanics.

It has been shown (33) that the choice of the coefficient $\beta$ is also critical in efficiency of the algorithm. If $\beta/y_\circ$ is larger than one, then a high diversity is imposed in the system and a more efficient exploration of the phase space is achieved. However, the cost of this choice is that algorithm becomes computationally more expensive, because of slow convergence of Eq. (7). Therefore, a good choice for $\beta$ would be such that $\beta/y_o$ is close to one. In addition, the value of $\beta/y_o$ depends on $P_o$. In Figure 1A, we have plotted the calculated $1/(\alpha_{t+1}|r_t|)$ or $1/y_o$ for different values of $P_o$ in the range 0.5 to one. It can be seen that for a value of $P_o = 0.75$, an appropriate choice of $\beta$ would be 0.5 to guarantee the convergence of Eq. (7).

Larger values of $P_o$ will guarantee on average faster convergence of the swarm particles (as indicated from Figure 1A), but the volume of the sampled space with be smaller, and thus not all the phase space point will be visited. For an efficient algorithm, avoiding premature convergences it is required that the convergence time length to be moderate. This is in analogy with a Monte Carlo acceptance transition probability; it should not be too small because large and not physical displacement will be accepted, but on the other hand, it should not be too large because the displacement will be small and the particles will not be able to visit long distant regions from the initial state. From the mathematical point of view, the choice of $P_0$ will determine the value of $y_0$, which enters into the expression for $1/\alpha_{t+1}$ (see Eq. (12)). This, on the other hand, will determine the width, $w = g\beta_t/y_0$, of the Gaussian distribution, which defines the diversity of the swarm particles. As it can be seen, this value is obtained up to constant $g$, which is fixed here to 0.5. Thus, tuning the value of constant $g$ around 0.5, but to be always less than one, will justify any value of $P_0$ around 0.75 chosen in this study.

It has been suggested elsewhere (37) (see also discussion in Ref. (14)) that the value of parameter $w$ has to be less than 1.7 in order to guarantee the convergence of the particles.

A sinusoidal expression such as $\beta = y_0 + y_0|A\,sin(\omega t)|$ has shown to be an improvement of the performance (37; 38) by means of the stochastic simulation. Here we choose the same mathematical form using several different values of $A$ in the range from 0.01 to 2.0, and $\omega = 0.1$.

For the q-GSQPO algorithm the situation is slightly different. The solution $y_o^{(q)}$ depends on both, $q$ and $P_o$. In Figure 1B we have calculated $1/y_o^{(q)}$ versus $P_o$ for different $q$ taken from a geometrical series in the range one to two. It can be seen, at $P_o = 0.75$, the value of $1/y_o^{(q)}$ lie in the range 1.0 to 1.5 for the values of $q$ depicted in Figure 1B.

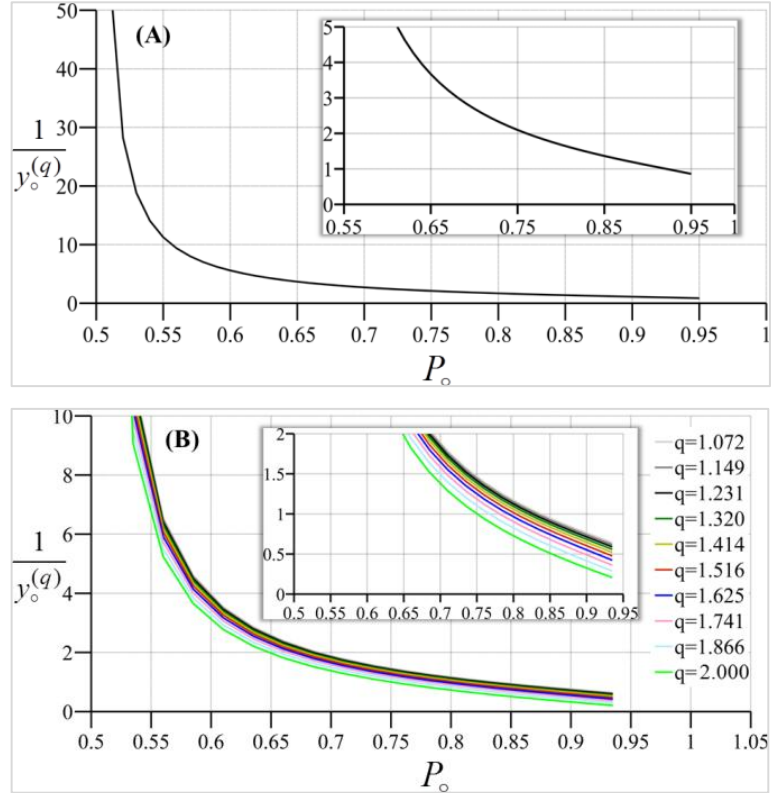

Figure 1: The inverse of the solution, $1/y_o$ (or $1/y_o^{(q)}$) versus $P_o$ for (A) GSQPO algorithm and (B) for q-GSQPO algorithm for different values of parameter $q$ taken from a geometrical series.

An exact numerical solution $y_o^{(q)}$ versus $q$ at $P_o = 0.75$ is presented in Figure 2. Also for q-GSQPO algorithm we found that a sinusoidal function of $\beta^{(q)}$, $\beta^{(q)} = y_o^{(q)} + y_o^{(q)} |A \sin(\omega t)|$, with the same value for $\omega$, yield better performance of the algorithm.

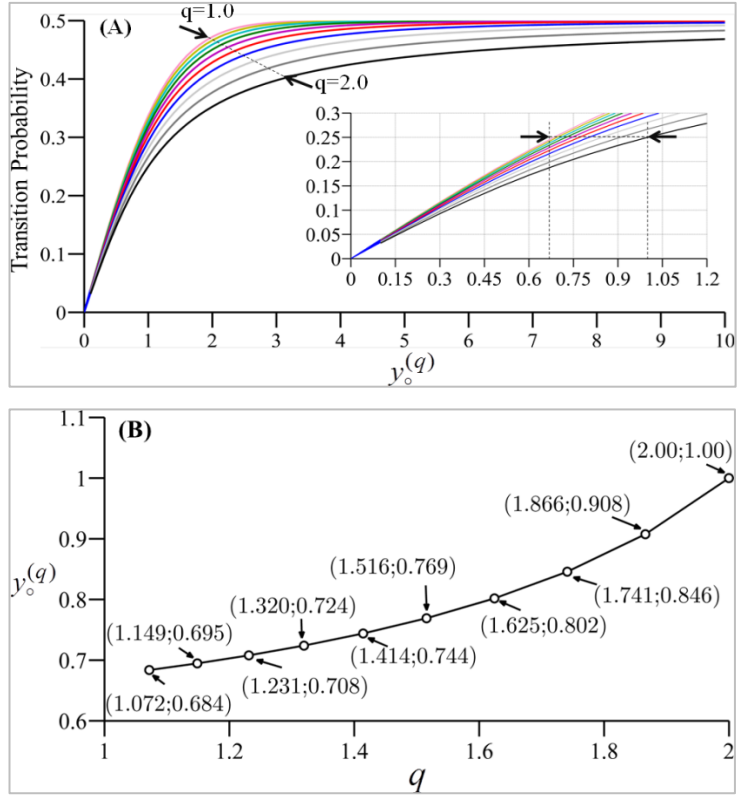

Figure 2: (A) Transition probability versus $y_\circ$ for different values of parameter $q$ in the range 1 to 2 taken as a geometric series. (B) Pairs ($q$, $y_\circ^{(q)}$) as solution of the Eq. (20) for $P_\circ = 0.75$.

It is found by the results of stochastic simulation that for $w$ less than some value 1.7, the particles will converge to some point, and when $w$ is larger than 1.8, they will diverge (37; 38; 16). The same results were also found in our study (data not shown). We chose as a threshold value of $w$ to be the value $w_0 = 1.71$, i.e., the particles will converge if $w \leq w_0$, otherwise they will diverge. There exist two methods of $w$: the linear-decreasing method and adaptive mechanism of parameter selection, where the later one controls the parameter on the global level and found to overcome the problem of premature convergence. Here, we also follow the adaptive selection method, but with slightly different form, which by empirical evaluations is found to perform better.

The main reason for premature convergence in the PSO algorithms is the low diversity of the particles (27). Some diversity control methods have been proposed in SQPO to enhance the ability of algorithm to escape the local minima (19; 17; 18). In this paper, diversity control of the swarm in the q-GSQPO is managed by the q-Gaussian distribution. Since q-Gaussian has longer tail distribution compare to Gaussian, this increases the diversity of the swarm and hence overcoming the problem of premature convergence. Thus, the method enhances the ability to escape the local minima.

### III. Analyzing the results

In order to estimate the efficiency of the algorithm, we calculated the failure rate (in percentage) of finding the global minimum out of $N_{runs}$ simulation runs. Each simulation run was stopped after a convergence criteria was satisfied, such that $\Delta < 10^{-5}$, with $\Delta$ being the diversity, given by

$$\Delta = \frac{1}{N} \sum_{i=1}^{N} \left| \vec{X}_i^{LBest}(t) - \vec{M}^{Best}(t) \right| \tag{22}$$

The number of swarm particles was $N_p = 5$. In addition, we also calculated the average number of iterations for the algorithm to convergence, defined as

$$\langle N_{iter} \rangle = \frac{1}{N_{runs}} \sum_{k=1}^{N_{runs}} N_{k,iter} \tag{23}$$

where $N_{k,iter}$ is the number of iterations of the $k$-th run.

### IV. Results and Discussion

We considered the motion in the $d$-dimensional potential hyper-surfaces determined by the mathematical functions summarized in **Table 1**.

The total number of runs was $N_{runs} = 10000$. To compare the algorithms, we plotted failure rate versus $\langle N_{iter} \rangle$ for each algorithm as shown in Error! Reference source not found. for the Griewank, Rastrigin and Ackley functions. In the case of q-GSQPO algorithm, we have shown results for three different values of parameter $q$: $q = 1.072, 1.32$, and $1.625$. The results are obtained by changing the value of parameter $A$ in the range $0.01 \div 0.45$. For large values of $A$, failure rate decreases and $\langle N_{iter} \rangle$ increases.

Table 1: Benchmark functions used for comparisons in this study.

| Mathematical functions | Reference | 2D Plot |
|---|---|---|
| $f_1(x_1, x_2, \cdots, x_d) = \dfrac{1}{400} \sum_{i=1}^{d} x_i^2 - \prod_{i=1}^{d} \cos(x_i / \sqrt{i}) + 1,$ $-6\pi \le x_i \le 6\pi$ | Griewank function (39) $f_1^*(x_i^*) = 0,$ $x_i^* = 0$ $i = 1, 2, \cdots, d$ | 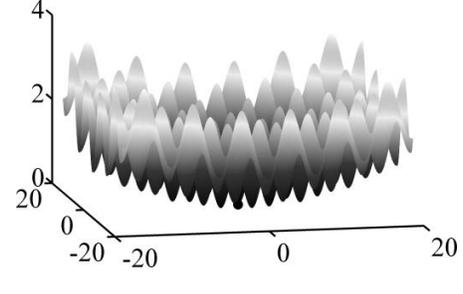 |
| $f_2(x_1, x_2, \cdots, x_d) = \sum_{i=1}^{d} \left(10 + x_i^2 - 10 \cdot \cos(2\pi x_i)\right),$ $-6\pi \le x_i \le 6\pi$ | Rastrigin function (40) $f_2^*(x_i^*) = 0,$ $x_i^* = 0$ $i = 1, 2, \cdots, d$ | 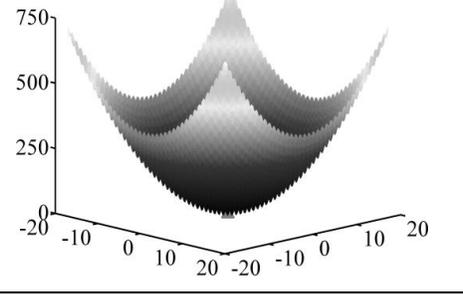 |
| $f_3(x_1, x_2, \cdots, x_d) = -20 \exp\left(-0.2 \sqrt{\dfrac{1}{d} \sum_{i=1}^{d} x_i^2}\right) - \exp\left(\dfrac{1}{d} \sum_{i=1}^{d} \cos(2\pi x_i)\right) + 20 + e,$ $-6\pi \le x_i \le 6\pi$ | Ackley function (41) $f_3^*(x_i^*) = 0,$ $x_i^* = 0$ $i = 1, 2, \cdots, d$ | 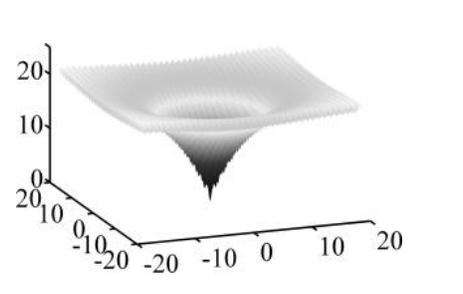 |

The interpolated results from the values obtained for $A$ in the above range are plotted. Our results indicate that failure rate increases with increasing the average iteration number for both algorithms. Most importantly, the q-GSQPO algorithm outperforms the GSQPO, in that; the failure rate is always smaller for the same $\langle N_{iter} \rangle$. In addition, we can see (Figure 3) that with increasing $q$ failure rate decreases for the same average number of iterations. Our results are consistent for all three benchmark functions (see Figure 3A, Figure 3B and Figure 3C).

An explanation of these findings is that heavy tail distributions yield a higher diversity of the swarm, and hence the particles visit long distant regions from the initial point of the search. This improvement in exploration of the search space will allow the particles to overcome the barriers that exist between the local minima on the hyper surface and so increase the efficiency of the converging to the global minimum.

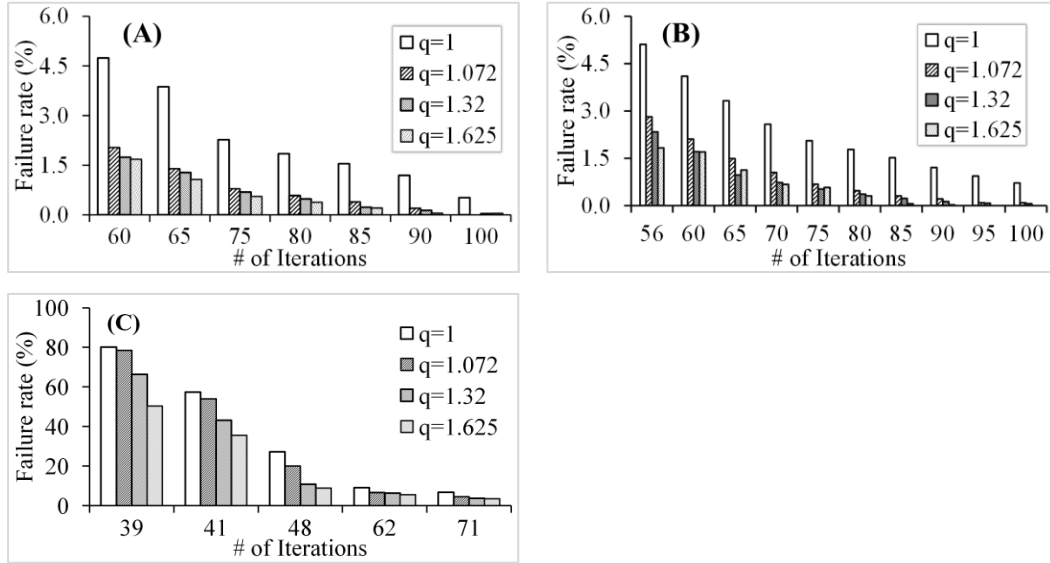

Figure 3: Failure rate (in percentage) versus number of the iterations for GSQPO ($q=1$) and q-GSQPO algorithms for $q=1.072$, $q=1.32$, and $q=1.625$. (A) Griewank function and (B) Rastrigin function and (C) Ackley function.

We also calculated the average best score versus the parameter $A$ for Rastrigin, Griewank and Ackley surfaces for three different dimensions 5, 10 and 50. Results are shown graphically in Figure 4 for d = 5, $N_p = 50$ and $N_{runs} = 2000$; (Middle) for d = 10, $N_p = 100$, and $N_{runs} = 2000$; (Bottom) for d = 50, $N_p = 500$, and $N_{runs} = 1000$. The values of parameter q are shown in the figures; A was taken in the range from 0.01 to 2.2. Our data show that for both algorithms, the average best score decreases as $A$ increases, because the width of the distributions, either Gaussian or q-Gaussian, increases allowing the swarm to sample long distant regions. Comparing the two algorithms, q-GSQPO converges faster to the global minimum, as also indicated from the values of the average best scores (see Figure 4). In addition, we can see that this is true also for higher dimensions, such as 50, for all three functions considered in our study.

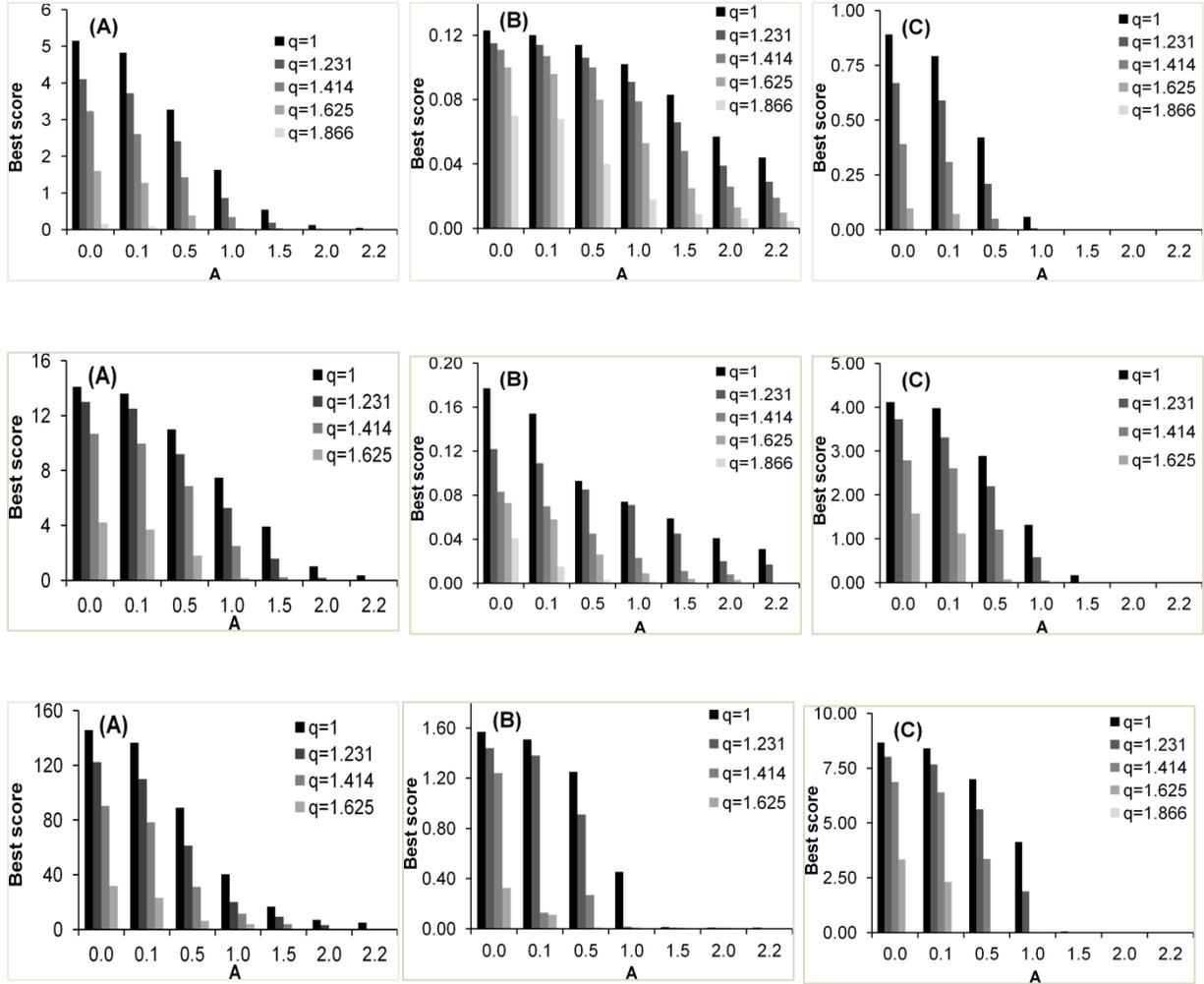

Figure 4: Average best score solution versus the value of A for (A) Rastrigin; (B) Griewank; and (C) Ackley functions in three different dimensions: (Top) $d = 5$, $N_p = 50$ and $N_{runs} = 2000$; (Middle) $d = 10$, $N_p = 100$, and $N_{runs} = 2000$; (Bottom) $d = 50$, $N_p = 500$, and $N_{runs} = 500$.

As we also mentioned before, the main merit of the q-GSQPSO algorithms is the efficiency of the phase space sampling. That is, this algorithm is more efficient avoiding the premature convergences by increasing the swarm diversity. To show that the new algorithm is more efficient in keeping high diversity of the swarm we also calculated the diversity versus number of iterations for both algorithms. We fixed the dimensionality of the space to 10, and we averaged out over 100 runs. Results are shown in Figure 5 for the three functions using 50 particles. Our data show that diversity according to the q-GSQPO algorithm decreases slower with number of iterations compared with GSQPO algorithm for all three functions. These results suggest that q-GSQPO explores better the phase space, since the swarm particles search a larger volume of the space. For clarity we are also showing the diversity up to longer number of iterations (see Figure 5 (Bottom)).

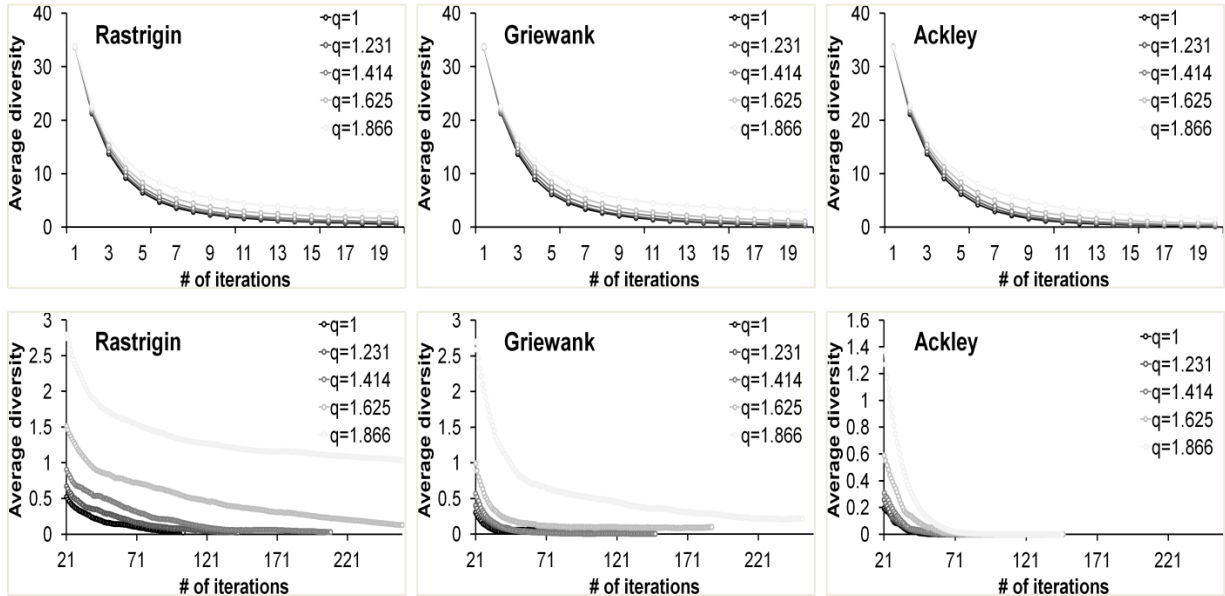

Figure 5: Average diversity versus the value number of iterations for Rastrigin, Griewank and Ackley 10 dimensional functions: (Top) for iterations from 1 to 20; (Bottom) for iterations from 20 to 250. Number of particles was $N_p = 50$ and $N_{runs} = 100$.

We also calculated the elapsed CPU times for both algorithms, and plotted results for different values of the parameter $A$, in the range 0.01 to 2.2, in Figure 6. The dimension was $d = 5$, and the number of particles $N_p = 50$. The results were averaged out of 1000 runs. The data are shown for three benchmark functions studied here. Since the CPU time is machine dependent, we normalized the CPU time of q-GSQPO algorithm, for different values of $q$, at the CPU time of GSQPO algorithm ($q = 1$). CPU time can be seen to dependent on the parameters $A$ and $q$. With increasing A (*i.e.*, the width of the swarm distribution function), in particular for large value of q, the CPU time of q-GSQPO algorithm increases compare to the CPU time of GSQPO algorithm in the ranges from 1.5 to 3. This is observed in particular for Ackley function, while for the other two functions CPU time for both algorithms is comparable. Another finding of our study is that for values of q up to

$$q = 1 + 1/\sqrt{d}$$

both algorithms have comparable CPU time. We also compared the CPU time for iteration (data not shown here) and we found that both algorithms perform at the same CPU time. This result is expected, since the algorithms differ from each other only on the distribution $F(u)$: for q-GSQPO algorithm this is q-Gaussian distribution and for GSQPO, it is Gaussian distribution. From the computation point of view, q-Gaussian random number and Gaussian random number are generated using equivalent amount of computational efforts.

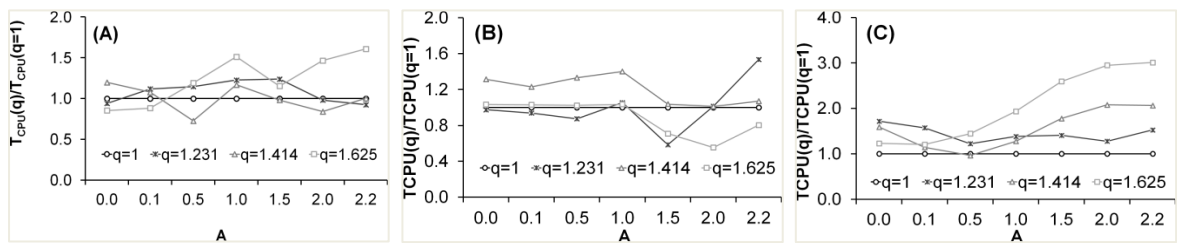

Figure 6: Average CPU time normalized for the CPU time at $q = 1$ versus the value parameter B for (A) Rastrigin, (B) Griewank and (C) Ackley 5 dimensional functions. Number of particles was $N_p = 50$ and $N_{runs} = 1000$.

## V.    Conclusions

A framework for q-Gaussian Quantum-like Swarm Particle Optimization (q-GSQPO) suitable for determining the global minimum of the potential energy function was proposed. Through choosing attractive potential field, such that corresponds to a q-Gaussian probability distribution function, an algorithm was obtained by solving the corresponding Schrödinger equation.

Considering prediction power of the algorithm as a function of the number of failures and the number of iterations, as common requirements in every evolutionary computing algorithm, we found that the q-GSQPO algorithm contains only one control parameter and provide a satisfactory convergence to the desired global optimal solution.

Furthermore, the heavy tails probability distribution allows retaining a high diversity of the swarm particles avoiding the premature convergences to local minimum.

In addition, the comparisons of the elapsed CPU time, indicate that restricting the maximum value of $q$ to $1 + 1/\sqrt{d}$, make the computational efforts of q-GSQPO and GSQPO algorithms comparable. We think that limiting the highest values of $q$ has a physical consequence. It is clear that increasing $q$, we increase the number of particles with small probabilities, but non-vanishing, which explore regions far from the origin, and hence slowing down the convergence of the algorithm. On the other hand, relatively small values of $q$ lead to a tighter distribution around the origin, which mean more particles are likely to be close to the origin, and thus faster the convergence. Therefore, the value of $q = 1 + 1/\sqrt{d}$ can be used as a critical value for the q-Gaussian probability distribution to retain the similarity in shape with a Gaussian distribution, but with longer tails.

## VI. Acknowledgements

The author would like to acknowledge the International Balkan University for the support.